\documentclass{article}

\usepackage[preprint]{nips_2018}

\usepackage[utf8]{inputenc} 
\usepackage[T1]{fontenc}    
\usepackage{hyperref}       
\usepackage{url}            
\usepackage{booktabs}       
\usepackage{amsfonts}       
\usepackage{nicefrac}       
\usepackage{microtype}      

\usepackage{amsmath}
\usepackage{tablefootnote}
\usepackage{tikz}
\usepackage{pgfplots}

\usetikzlibrary{positioning,shapes}

\usepackage{ifthen}
\usepackage{xcolor}

\newcommand*{\todo}[2][]{\textcolor{red}{[\textbf{\ifthenelse{\equal{#1}{}}{TODO}{TODO(#1)}}: #2]}}

\def\etal.{et\penalty50\ al.}

\newcommand*\samethanks[1][\value{footnote}]{\footnotemark[#1]}

\DeclareMathOperator*{\argmax}{arg\,max}

\newcommand{\footlabel}[2]{%
    \addtocounter{footnote}{1}%
    \footnotetext[\thefootnote]{%
        \addtocounter{footnote}{-1}%
        \refstepcounter{footnote}\label{#1}%
        #2%
    }%
    $^{\ref{#1}}$%
}

\title{Evaluating and Understanding the Robustness of Adversarial Logit Pairing}

\author{
    Logan Engstrom\thanks{Equal contribution}\hspace{2em} Andrew Ilyas\samethanks\hspace{2em} Anish Athalye\samethanks \\
    Massachusetts Institute of Technology \\
    \texttt{\{engstrom,ailyas,aathalye\}@mit.edu}
}

\begin{document}

\maketitle

\begin{abstract}
    We evaluate the robustness of Adversarial Logit Pairing, a recently proposed
defense against adversarial examples. We find that a network trained with
Adversarial Logit Pairing achieves 0.6\% correct classification rate under
targeted adversarial attack, the threat model in which
the defense is considered. We provide a brief overview of the defense and the
threat models/claims considered, as well as a discussion of the methodology and
results of our attack. Our results offer insights into the reasons underlying the
vulnerability of ALP to adversarial attack, and are of general interest
in evaluating and understanding adversarial defenses.

\end{abstract}

\section{Contributions}
For summary, the contributions of this note are as follows:
\begin{enumerate}
    \item \textbf{Robustness}: Under the white-box targeted attack threat
model specified in~\citet{alp}, we upper bound the correct classification
rate of the defense to
	\textbf{0.6\%} (Table~\ref{tab:claims}). We also perform targeted
	and untargeted attacks and show that the attacker can reach success
	rates of 98.6\% and 99.9\% respectively
(Figures~\ref{fig:eps-success},~\ref{fig:eps-accuracy}).
    \item \textbf{Formulation}: We analyze the ALP loss function and
	contrast it to that of~\citet{madry2018towards}, pointing out
several differences from the robust optimization objective
(Section~\ref{sec:obj}).
    \item \textbf{Loss landscape}: We analyze the loss landscape induced by
	ALP by visualizing loss landscapes and
	adversarial attack trajectories (Section~\ref{sec:emp}).
\end{enumerate}

Furthermore, we suggest the experiments conducted in the analysis of ALP as
another evaluation method for adversarial defenses.

\section{Introduction}
\label{sec:introduction}

Neural networks and machine learning models in general are known to be
susceptible to adversarial examples, or slightly perturbed inputs that
induce specific and unintended
behaviour~\citep{szegedy2013intriguing,biggio2013evasion}.
Defenses against these adversarial attacks are of great significance
and value. Unfortunately, many proposed defenses have had their claims invalidated by new attacks within their corresponding threat models~\citep{carlini2016distillation,
he2017ensembles,carlini2017adversarial,carlini2017magnet,obfuscated,uesato2018risk,cvprbreaks}.
A notable defense has been that
of~\citet{madry2018towards},
which proposes a ``robust optimization''-based view of defense against
adversarial examples, in which the defender tries to find parameters $\theta^*$
minimizing the following objective:

\begin{equation}
\label{eq:madry}
\min_{\theta} \mathbb{E}_{(x,y)\sim\mathcal{D}}\left[\max_{\delta \in
S} L(\theta, x+\delta, y)\right].
\end{equation}

Here, $L$ is a prespecified loss function, $\mathcal{D}$ is the labeled
data distribution, and $\mathcal{S}$ is the set of admissible adversarial
perturbations (specified by a threat model). In practice, the defense is
implemented through adversarial training,
where adversarial examples are generated during the training process and
used as inputs. The resulting classifiers have been empirically evaluated to
offer increased robustness to adversarial examples on the CIFAR-10 and
MNIST datasets under small $\ell_\infty$ perturbations.

In~\citet{alp}, the authors claim that the defense of \citet{madry2018towards} is ineffective when scaled to an ImageNet~\citep{deng2009imagenet} classifier, and propose a new defense: Adversarial Logit Pairing (ALP). In the ALP defense, a classifier is trained with a training objective that enforces similarity between the
model's \textit{logit} activations on unperturbed and
adversarial versions of the same image. The loss additionally has a term meant to maintain accuracy on
the original training set.
\begin{align*}
    \min_{\theta} \mathbb{E}_{(x, y)\sim\mathcal{D}}\left[L(\theta, x, y) + \lambda D\left(f(\theta,
    x), f(\theta, x+\delta^{*})\right)\right] \\ \text{where } \delta^{*} =
    \argmax_{\delta\in\mathcal{S}} L(\theta, x+\delta, y),
\end{align*}
Here, $D$ is a distance function, $f$ is a function mapping parameters and
inputs to logits (via the given network), $\lambda$ is a
hyperparameter, and the rest of the notation is as in~\eqref{eq:madry}.
This objective is intended to promote ``better internal representations of
the data''~\citep{alp} by providing an extra regularization term.
In the following sections, we show that
ALP can be circumvented using Projected Gradient Descent (PGD) based attacks.

\subsection{Setup details}
\label{sec:setup}

We analyze Adversarial Logit Pairing as implemented by the
authors~\footlabel{footnote:alp}{\scriptsize\url{https://github.com/tensorflow/models/tree/master/research/adversarial_logit_pairing}}.
We use the ``models pre-trained on
ImageNet'' from the
code release to evaluate the claims of~\citet{alp}. Via private correspondence, the authors acknowledged our result but
stated that the results in~\citet{alp} were generated with different,
unreleased models not included in the official code release.

Our evaluation code is publicly
available.~\footnote{\scriptsize\url{https://github.com/labsix/adversarial-logit-pairing-analysis}}.

\section{Threat model and claims}
\label{sec:background}

\begin{table}[htbp]
\begin{center}
    \caption{{\small The claimed robustness of Adversarial Logit Pairing against
    targeted attacks on ImageNet, from \citet{alp}, compared to the lower
    bound on attacker success rate from this work. Attacker success rate in
this case represents the percentage of times an attacker successfully induces
the adversarial target class, whereas accuracy measures the percentage of times
the classifier outputs the \textit{correct} class. \vspace{1em}}}
\label{tab:claims}
\begin{tabular}{lccc}
    \toprule
    \textbf{Source} & \citet{alp} & this work & this work \\
    \textbf{Defense $(\epsilon = 16/255)$} & \textbf{Claimed Accuracy}
					  & \textbf{Defense
Accuracy\tablefootnote{We calculate this as in~\cite{alp}, i.e.
correct classification rate under targeted adversarial attack.}}
					  & \textbf{Attacker Success} \\
    \midrule
    \citet{madry2018towards} & 1.5\% & -- & -- \\
    \citet{alp} & 27.9\%\tablefootnote{As noted in \S\ref{sec:setup}, via
    private correspondence, the authors state that
unreleased models were used to generate the results in~\citet{alp}. The
authors are currently investigating these models; for the sake of comparison, we give the claim
from~\citet{alp} here.} & \textbf{0.6\%} & \textbf{98.6\%} \\ \bottomrule
\end{tabular}
\end{center}
\end{table}

ALP is claimed secure under a variety of white-box and black-box threat models;
in this work, we consider the \textit{white-box} threat model, where an
attacker has full access to the weights and parameters of the model being
attacked. Specifically, we consider an Residual
Network ALP-trained on the ImageNet dataset, where ALP
is claimed to achieve state-of-the-art accuracies in this setting under an
$\ell_\infty$ perturbation bound of $16/255$, as shown in
Table~\ref{tab:claims}. The defense is originally evaluated against
targeted adversarial attacks, and thus Table~\ref{tab:claims} refers to the
attacker success rate on targeted adversarial attacks. For completeness, we
also perform a brief analysis on untargeted attacks to show lack of
robustness (Figure~\ref{fig:eps-accuracy}), but do not consider this in the
context of the proposed threat model or claims.

\paragraph{Adversary objective.}
When evaluating attacks, an attack that can produce targeted adversarial
examples is \textit{stronger} than an attack that can only produce untargeted
adversarial examples. On the other hand, a defense that is only robust against
targeted adversarial examples (e.g. with random target classes) is
\textit{weaker} than a defense that is robust against untargeted adversarial
examples. The ALP paper only attempts to show robustness to targeted
adversarial examples.

\section{Evaluation}

\begin{figure}
    \begin{minipage}[t]{.45\textwidth}
    \center
    \begin{tikzpicture}[scale=0.8]
    \newcommand{\labelpos}{50}
    \begin{axis}[
            title=\textbf{Attack success rate (lower is better)},
            xlabel=$\epsilon$,
            ylabel=Attack success rate (\%),
            xmin=0,
            xmax=17,
            ymin=0,
            ymax=100,
            legend style={at={(0.5,0.03)},anchor=south},
        ]
        \addplot[color=blue,mark=+] file {data/alp-targeted.txt};
        \addplot[color=red,mark=x] file {data/base-targeted.txt};
        \legend{ALP,Baseline}

        \draw[color=black,thick,dashed] (axis cs:16, 0)
                             -- (axis cs:16, 100);
        \node (mid) at (axis cs:16, \labelpos) {};
        \node[rotate=90,left = 0 of mid,anchor = south] (label) {considered threat model};
    \end{axis}
\end{tikzpicture}
        \caption{Comparison of ALP-trained model with baseline model under \textbf{targeted}
        adversarial perturbations (with random labels) bounded by varying
        $\epsilon$ from $0$ to $16/255$. Our attack reaches 98.6\% success rate
        (and 0.6\% correct classification rate) at $\epsilon = 16/255$. 
        }
    \label{fig:eps-success}
    \end{minipage}
    \hspace{0.1\textwidth}
    \begin{minipage}[t]{.45\textwidth}
    \center
    \begin{tikzpicture}[scale=0.8]
    \newcommand{\labelpos}{50}
    \begin{axis}[
            title=\textbf{Accuracy (higher is better)},
            xlabel=$\epsilon$,
            ylabel=Accuracy (\%),
            xmin=0,
            xmax=17,
            ymin=0,
            ymax=100,
            legend pos=north west,
        ]
        \addplot[color=blue,mark=+] file {data/alp-untargeted.txt};
        \addplot[color=red,mark=x] file {data/base-untargeted.txt};
        \legend{ALP,Baseline}

        \draw[color=black,thick,dashed] (axis cs:16, 0)
                             -- (axis cs:16, 100);
        \node (mid) at (axis cs:16, \labelpos) {};
        \node[rotate=90,left = 0 of mid,anchor = south] (label) {considered threat model};
    \end{axis}
\end{tikzpicture}
    \caption{Comparison of ALP-trained model with baseline model under
        \textbf{untargeted} adversarial perturbations bounded by varying $\epsilon$ from $0$
    to $16/255$. The ALP-trained model achieves 0.1\% accuracy at $\epsilon =
    16/255$.}
    \label{fig:eps-accuracy}
    \end{minipage}
\end{figure}
\label{sec:breaks}

\subsection{Analyzing the defense objective}
\label{sec:obj}
Adversarial Logit Pairing is proposed as an
augmentation of adversarial training, which itself is meant to
approximate the robust optimization approach outlined in
Equation~\ref{eq:madry}. The paper claims that by adding a
``regularizer'' to the adversarial training objective, better results on
high-dimensional datasets can be achieved. In this section we
outline several conceptual differences between ALP and the
robust optimization perspective offered by~\citet{madry2018towards}.

\paragraph{Training on natural vs. adversarial images.} A key part in the
formulation of the robust optimization objective is that minimization
with respect to $\theta$ is done over the inputs that have been crafted by
the $\max$ player; $\theta$ is not minimized with respect to any ``natural''
$x \sim \mathcal{D}$. In the ALP formulation, on the other hand,
regularization is applied to the loss on \textit{clean} data $L(\theta, x,
y)$. This fundamentally changes the optimization objective from the defense
of~\citet{madry2018towards}.

\paragraph{Generating targeted adversarial examples.} A notable
implementation decision given in~\citet{alp} is to generate targeted
adversarial examples during the training process. This again deviates from
the robust optimization-inspired saddle point formulation for adversarial
training, as the inner maximization player no longer maximizes $L(\theta,
x+\delta, y)$, but rather minimizes $L(\theta, x+\delta, y_{adv})$ for
another class $y_{adv}$. Note that although~\citet{obfuscated}
recommends that \textit{attacks} on ImageNet classifiers be evaluated in the
targeted threat model (which is noted in~\citep{alp} in justifying this
implementation choice), this recommendation does not extend to adversarial
training or empirically showing that a defense is secure (a defense that is only robust to targeted
attacks is \textit{weaker} than one robust to untargeted attacks).

\subsection{Analyzing empirical robustness}
\label{sec:emp}
Empirical evaluations give upper bounds for the robustness of a defense on test
data. Evaluations done with weak attacks can be seen as giving loose bounds,
while evaluations done with stronger attacks give tighter bounds of true
adversarial risk~\citep{uesato2018risk}. We find that the robustness of ALP as a defense to adversarial examples is
significantly lower than claimed in \citet{alp}.

\paragraph{Attack procedure.}
We originally used the evaluation code provided by the ALP authors and
found that setting the number of steps in the PGD attack to 100 from the
default of 20 significantly degrades accuracy. For ease of use we
reimplemented a standard PGD attack, which we ran for up to 1000 steps or
until convergence. We evaluate both untargeted attacks and targeted attacks
with random targets, measuring model accuracy on the former and adversary
success rate (percentage of data points classified as the target class) for
the latter.

\paragraph{Empirical robustness.} We establish tighter upper bounds on adversarial
robustness for both the ALP trained classifier and the baseline (naturally trained)
classifier with our attack. Our results, with a full
curve of $\epsilon$ (allowed perturbation) vs attack success rate, are
summarized in Figure~\ref{fig:eps-success}. In the threat model with
$\epsilon=16$ our attack achieves a 98.6\% success rate and reduces the
accuracy (percentage of correctly classified examples perturbed by the
targeted attack) of the classifier to \textbf{0.6\%}.

Figure~\ref{fig:eps-accuracy} shows that untargeted attacks gives similar results: the ALP-trained model achieves 0.1\% accuracy at $\epsilon = 16/255$.

\begin{figure}
    \center
    \begin{center}
        \makebox[\textwidth]{\includegraphics[width=\textwidth]{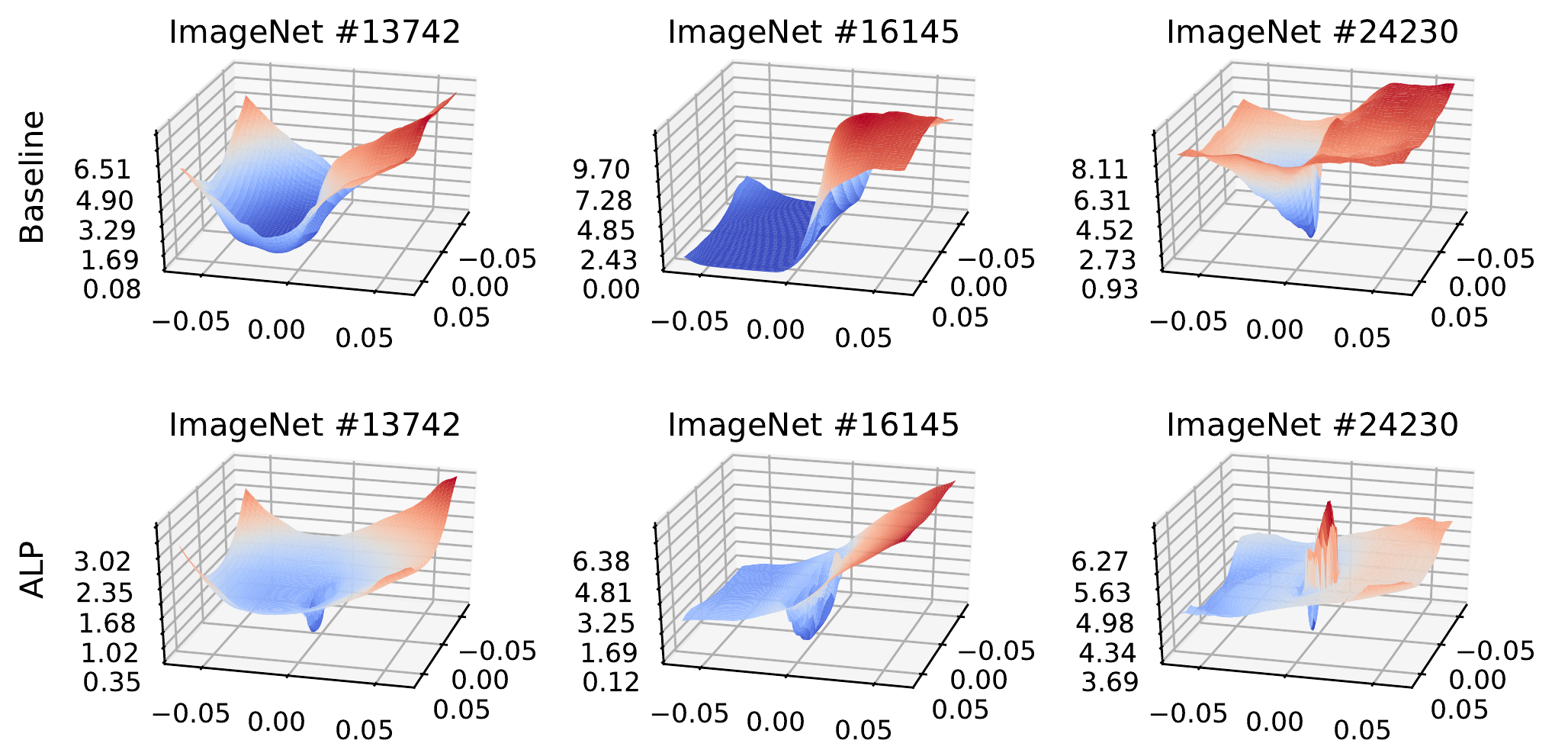}}
    \end{center}

    \caption{Comparison of loss landscapes of ALP-trained model and baseline model. Loss plots are generated by varying the input to the models, starting from an original input image chosen from the test set. We see that ALP sometimes induces decreased loss near the input locally, and gives a ``bumpier'' optimization landscape. The z axis represents the loss. If $\hat{x}$ is the original input, then we plot the loss varying along the space determined by two vectors: $r_1 = \text{sign}(\nabla_x f(\hat{x}))$ and $r_2 \sim \text{Rademacher(0.5)}$. We thus plot the following function: $z = \text{loss}(x \cdot r_1 + y \cdot r_2)$. The classifier here takes inputs scaled to $[0,1]$.}
    \label{fig:3d-plots}
\end{figure}

\paragraph{Loss landscapes.} We plot loss landscapes around test data points in Figure~\ref{fig:3d-plots}. We vary the input along a linear space defined by the sign of the gradient and a random Rademacher vector, where the x and y axes represent the magnitude of the perturbation added in each direction and the z axis represents the loss. The plots provide evidence that ALP sometimes induces a ``bumpier,'' depressed loss landscape tightly around the input points.

\paragraph{Attack convergence.}

As suggested by analysis of the loss surface, the optimization landscape of the
ALP-trained network is less amenable to gradient descent. Examining, for a
single data point, the loss over steps of gradient descent in targeted
(Figure~\ref{fig:loss-targeted}) and untargeted
(Figure~\ref{fig:loss-untargeted}) attacks, we observe that the attack on the
ALP-trained network takes more steps of gradient descent.

This was generally true over all data points. The attack on the ALP-trained
network required more steps of gradient descent to converge, but
robustness had not increased (e.g. at $\epsilon = 16/255$, both networks have
roughly $0\%$ accuracy).

\begin{figure}
    \begin{minipage}[t]{.45\textwidth}
    \center
    \begin{tikzpicture}[scale=0.8]
    \newcommand{\alpfinish}{240}
    \newcommand{\basefinish}{17}
    \begin{axis}[
            xlabel=PGD Step,
            ylabel=$- \log P(\text{target class})$,
            legend style={at={(0.5,0.97)},anchor=north},
        ]
        \addplot[color=blue,mark=+] file {data/alp_eps16_loss_targeted_3.txt};
        \addplot[color=red,mark=x] file {data/base_eps16_loss_targeted_3.txt};
        \legend{ALP,Baseline}

        \draw[color=blue, dashed] (axis cs:\alpfinish, \pgfkeysvalueof{/pgfplots/ymin}) -- (axis cs:\alpfinish, \pgfkeysvalueof{/pgfplots/ymax});
        \draw[color=red, dashed] (axis cs:\basefinish, \pgfkeysvalueof{/pgfplots/ymin}) -- (axis cs:\basefinish, \pgfkeysvalueof{/pgfplots/ymax});
    \end{axis}
\end{tikzpicture}
    \caption{Comparison of targeted attack on ALP-trained model versus baseline
    model on a single data point, showing loss over PGD steps. Vertical lines
    denote the step at which the attack succeeded (in causing classification as
    the target class). The optimization process requires more gradient descent
    steps on the ALP model but still succeeds.}
    \label{fig:loss-targeted}
    \end{minipage}
    \hspace{0.1\textwidth}
    \begin{minipage}[t]{.45\textwidth}
    \center
    \begin{tikzpicture}[scale=0.8]
    \newcommand{\alpfinish}{5}
    \newcommand{\basefinish}{1}
    \begin{axis}[
            xlabel=PGD Step,
            ylabel=$- \log P(\text{true class})$,
            legend pos=south east,
        ]
        \addplot[color=blue,mark=+] file {data/alp_eps16_loss_untargeted_6.txt};
        \addplot[color=red,mark=x] file {data/base_eps16_loss_untargeted_6.txt};
        \legend{ALP,Baseline}

        \draw[color=blue, dashed] (axis cs:\alpfinish, \pgfkeysvalueof{/pgfplots/ymin}) -- (axis cs:\alpfinish, \pgfkeysvalueof{/pgfplots/ymax});
        \draw[color=red, dashed] (axis cs:\basefinish, \pgfkeysvalueof{/pgfplots/ymin}) -- (axis cs:\basefinish, \pgfkeysvalueof{/pgfplots/ymax});
    \end{axis}
\end{tikzpicture}
    \caption{Comparison of untargeted attack on ALP-trained model versus
    baseline model on a single data point, showing loss over PGD steps.
    Vertical lines denote the step at which the attack succeeded (in causing
    misclassification). The optimization process requires more gradient descent
    steps on the ALP model but still succeeds.}
    \label{fig:loss-untargeted}
    \end{minipage}
\end{figure}

\section{Conclusion}
\label{sec:conclusion}
In this work, we perform an evaluation of the robustness of the Adversarial
Logit Pairing defense (ALP) as proposed in~\citet{alp}, and show that it is not
robust under the considered threat model. We then study the formulation,
implementation, and loss landscape of ALP. The
evaluation methods we use are general and may help in enhancing evaluation
standards for adversarial defenses.

\section*{Acknowledgements}

We thank Harini Kannan, Alexey Kurakin, and Ian Goodfellow for releasing open-source code and pre-trained
models for Adversarial Logit Pairing.

{\small
\bibliographystyle{abbrvnat}
\bibliography{paper}
}

\end{document}